\documentclass[runningheads]{llncs}
\usepackage[T1]{fontenc}
\usepackage{algorithm}%
\usepackage{algorithmicx}%
\usepackage{algpseudocode}%
\usepackage{xcolor}
\usepackage{hyperref}
\usepackage{graphicx}
\usepackage{booktabs} 

\usepackage{multirow}
\begin{document}

\title{Converting Time Series Data to Numeric Representations Using Alphabetic Mapping and $k$-mer strategy}

\titlerunning{Converting Time Series Data to Numeric Representations}

\author{Sarwan Ali$^{1*}$\and
Tamkanat E Ali$^{2*}$\and 
Imdad Ullah Khan \inst{2}\and 
Murray Patterson \inst{1}
}
\authorrunning{S. Ali et al.}

\institute{$^1$Department of Computer Science, Georgia State University, Atlanta, Georgia, USA
\\
$^2$Department of Computer Science, Lahore University of Management Sciences, Lahore, Pakistan
\\
\email{sali85@student.gsu.edu, 20100159@lums.edu.pk, imdad.khan@lums.edu.pk, mpatterson30@gsu.edu}
\\
$^*$Equal Contribution
}

\maketitle    

\begin{abstract}
  In the realm of data analysis and bioinformatics, representing time series data in a manner akin to biological sequences offers a novel approach to leverage sequence analysis techniques. 
  Transforming time series signals into molecular sequence-type representations allows us to enhance pattern recognition by applying sophisticated sequence analysis techniques (e.g. $k$-mers based representation) developed in bioinformatics, uncovering hidden patterns and relationships in complex, non-linear time series data.
  This paper proposes a method to transform time series signals into biological/molecular sequence-type representations using a unique alphabetic mapping technique. 
  By generating 26 ranges corresponding to the 26 letters of the English alphabet, each value within the time series is mapped to a specific character based on its range. This conversion facilitates the application of sequence analysis algorithms, typically used in bioinformatics, to analyze time series data. We demonstrate the effectiveness of this approach by converting real-world time series signals into character sequences and performing sequence classification. The resulting sequences can be utilized for various sequence-based analysis techniques, offering a new perspective on time series data representation and analysis.
\end{abstract}



\section{Introduction}

Time series data is ubiquitous in various fields such as finance~\cite{kuo2021solving}, healthcare~\cite{ukil2021l1}, environmental monitoring~\cite{abilasha2022deep,zhang2017method}, and industrial processes~\cite{mehdiyev2017time}. Traditionally, analyzing time series data involves techniques like statistical modeling, machine learning, and signal processing. However, these methods often struggle with capturing complex, non-linear patterns inherent in many time series datasets~\cite{weerakody2021review}. In bioinformatics, sequence analysis has proven to be highly effective for studying biological sequences such as DNA, RNA, and proteins, uncovering patterns and relationships that are crucial for understanding biological functions and processes~\cite{ali2021k}. 
Furthermore, converting time series data to sequences can improve classification performance by utilizing effective sequence classification algorithms, which can be particularly useful in industrial monitoring to classify different machine states or detect early signs of equipment failure.
This paper explores the novel idea of leveraging these powerful sequence analysis techniques~\cite{ali2021spike2vec} for time series data by transforming time series signals into representations similar to biological sequences.

The concept of transforming time series data into sequence-based representations is not entirely new. Previous research has explored various symbolic representation methods, such as Symbolic Aggregate approXimation (SAX) and Piecewise Aggregate Approximation (PAA), to simplify and discretize time series data~\cite{ralanamahatana2005mining,yi2000fast,keogh2001dimensionality}. These methods, however, often lack the ability to fully capture the complexity and rich information contained in the original time series~\cite{chen2023joint}. Recent advances in representation learning and the success of $k$-mers-based representation in bioinformatics highlight the potential of adopting bioinformatics techniques for time series analysis~\cite{ali2022efficient}. The $k$-mers method, for example, segments sequences into overlapping substrings of length $k$, which can then be analyzed to uncover meaningful patterns and features~\cite{ali2021k}.

In this paper, we propose a method to transform time series signals into biological/molecular sequence-type representations using a unique alphabetic mapping technique. By dividing the range of time series values into $26$ distinct ranges, each corresponding to a letter of the English alphabet, we map each value in the time series to a specific character. This transformation results in a character sequence that retains the temporal ordering and relative magnitudes of the original time series data. The generated sequences can then be analyzed using sequence analysis algorithms from bioinformatics, such as $k$-mers-based analysis, which have been effectively used for identifying patterns in biological sequences.
To demonstrate the effectiveness of our proposed method, we conducted experiments using a real-world time series dataset originally used for human activity recognition~\cite{ali2023information}. We transformed these datasets into character sequences and applied sequence classification and evaluated their performance. The results show that our approach not only simplifies the representation of time series data but also enhances classification accuracy by leveraging the robust sequence analysis techniques from bioinformatics. 

The remainder of this paper is organized as follows: Section~\ref{sec_RW} discusses the previous literature.
Section~\ref{sec_PA} provides a detailed description of the proposed method for transforming time series signals into sequence-type representations. Section~\ref{sec_ES} outlines the experimental setup and datasets used for evaluation. Section~\ref{sec_RD} presents the results of our experiments and discusses the implications of our findings. Finally, Section~\ref{sec_conclusion} concludes the paper and suggests potential directions for future research

\section{Literature Review}\label{sec_RW}
The transformation of time series data into symbolic representations has been the subject of extensive research. Symbolic Aggregate approXimation (SAX) and Piecewise Aggregate Approximation (PAA) are among the most well-known methods for discretizing time series data. SAX reduces dimensionality and discretizes time series by mapping data points to a symbolic representation~\cite{ralanamahatana2005mining,yi2000fast}. PAA, on the other hand, divides the time series into equal-sized segments and calculates the mean value of each segment~\cite{keogh2001dimensionality}. While these methods simplify and make the data more manageable, they often fail to capture the intricate patterns and rich information present in complex time series data~\cite{chen2023joint}.

Recent advancements in representation learning have opened new avenues for time series analysis. Representation learning aims to automatically discover the representations or features required for a specific task. Methods such as $k$-mers-based representation, which originated in bioinformatics for analyzing biological sequences, have shown promising results in uncovering meaningful patterns in time series data~\cite{ali2022efficient,alipanahi2015predicting}. The $k$-mers approach involves segmenting sequences into overlapping substrings of length $k$, enabling the capture of local sequence information and facilitating more effective pattern recognition~\cite{ali2021k,greenberg2023analysis}.

In addition to traditional symbolic representation methods, more sophisticated techniques have been developed to enhance time series classification and analysis. One such technique is the use of convolutional neural networks (CNNs) and recurrent neural networks (RNNs) to learn representations directly from raw time series data. These models have demonstrated significant improvements in classification accuracy and the ability to handle the non-linear dynamics of time series~\cite{ismail2019deep}. However, these methods often require large amounts of labeled data and substantial computational resources, making them less practical for certain applications~\cite{ismail2020inceptiontime}.

The bioinformatics domain offers a rich set of sequence analysis tools and algorithms that can be leveraged for time series analysis. For example, sequence alignment algorithms, which are widely used for comparing DNA, RNA, and protein sequences, can be adapted to align and compare time series sequences~\cite{durbin1998biological}. Techniques such as hidden Markov models (HMMs) and dynamic time warping (DTW) have been successfully applied to time series data, demonstrating the potential of bioinformatics-inspired methods in this field~\cite{rabiner1989tutorial,berndt1994using}. The adoption of these techniques for time series analysis enables the discovery of complex temporal patterns and relationships that are often missed by traditional methods~\cite{jeong2011weighted}.

\section{Proposed Approach}\label{sec_PA}
\begin{figure}[h!]
    \centering
    \includegraphics[scale=0.043]{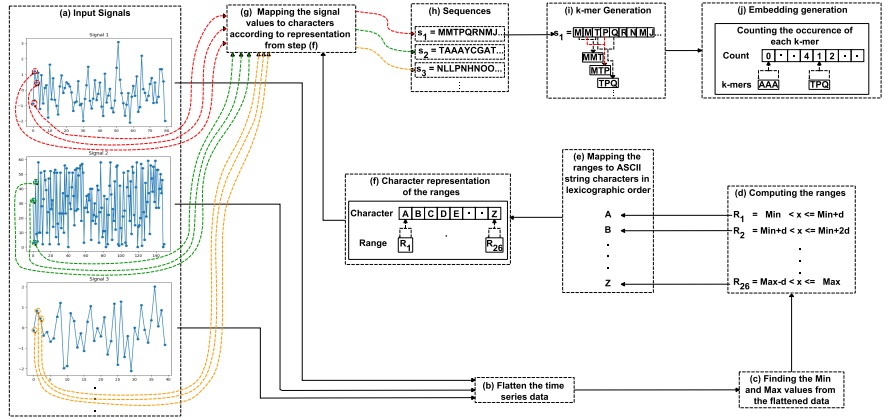}
    \caption{Flow diagram for the proposed approach}
    \label{proposed_approach}
\end{figure}

In this section, we present the details of our proposed method that generates numeric representation from time series data, the flow of our approach is shown in Figure~\ref{proposed_approach}. Given a set of Time Series Signal data as shown in step (a) of Figure~\ref{proposed_approach}, we generate sequence embedding using Algorithm~\ref{algo_embedding}. The first step as seen in line number 3 of Algorithm~\ref{algo_embedding} and step(b) of Fig~\ref{proposed_approach}, involves flattening the Time Series Signals data to assess and prepare the data for alphabetic mapping including finding the adequate range boundaries that divide the data into equal intervals in line number 4 of Algorithm~\ref{algo_embedding} using function'\textsc{ComputeRanges()}' presented in Algorithm~\ref{algo_range}. This function uses the flattened data to find the global maximum and minimum values in the time series data as seen in lines 2 and 3 of Algorithm~\ref{algo_range} and step (c) of Fig~\ref{proposed_approach}. In our case we choose 26 as the number of ranges in which we want the data to be divided which is in accordance with the total 26 Alphabets. This is followed by calculating the interval $d$ of each range using the maximum value, minimum value, and the total number of ranges using the equation in line number 5 of Algorithm~\ref{algo_range}. Using this interval value we compute the bounds of all the ranges in line 11 of Algorithm~\ref{algo_range} as depicted in step (d) of Figure~\ref{proposed_approach}.

Using these range boundaries and each signal from the original time series data we conduct the alphabetic mapping using Algorithm~\ref{algo_conversion} by calling the function '\textsc{Mapping()}' in line number 10 of Algorithm~\ref{algo_embedding}. As depicted by Algorithm~\ref{algo_conversion} and the colored lines linked to step (g) of figure~\ref{proposed_approach}, this function at a time takes one signal as input and checks each value of the signal according to Range bounds to see which range that value lies in as shown in line 7 of Algorithm~\ref{algo_conversion} and then based on the alphabet mapping rule observed in moving from step (d) to step (f) in Figure~\ref{proposed_approach}, each value belonging to a particular range is mapped to the respective alphabet in line 8 of Algorithm~\ref{algo_conversion}. As a result, we get a unique sequence of characters for each signal as depicted in step (h) of Figure~\ref{proposed_approach} which is further used as input to '\textsc{Computekmers()}' function in line number 11 of Algorithm~\ref{algo_embedding} that works according to Algorithm~\ref{algo_kmers} and extracts the $k$-mers from the sequence following the criteria shown in line 5 of Algorithm~\ref{algo_kmers} and step (i) of Fig~\ref{proposed_approach}. 

To finally generate the numeric embedding we use $k$-mers and their counts in each sequence as the numeric representation. For this, we represent each sequence as a spectrum that contains all possible $k$-mers that we get from line 7 of Algorithm~\ref{algo_embedding} and compute their respective counts representing the number of times a particular kmer occurs in a sequence as shown in line 16 of Algorithm~\ref{algo_embedding}. The spectrum of each sequence is further appended in line number 18 of Algorithm~\ref{algo_embedding} to form our final numeric representation of the time series data. These representations prove to be valuable for downstream tasks such as classification.

\begin{algorithm}[h!]
\begin{algorithmic}[1]
\Statex \textbf{Input: }\texttt{Set of Time Series Signals ($T$), Alphabets (A)}
\Statex \textbf{Output: }\texttt{Sequence Embedding ($\phi$)}
\Function{GenerateEmbedding}{$T$}
\State $\phi$$\gets$ [] \Comment{Initialize an empty list}
\State FlatData $\gets$ Flatten($T$) \Comment{Flatten the time series signal data}
\State RangeBounds $\gets$ \Call{ComputeRanges}{FlatData}

\State Num $\gets$ Length($T$) \Comment{Number of Signals in input data}
\State kmersize $\gets$ 3 \Comment{$k$-mer size}
\State \texttt{$Globalkmers \gets Combinations(A)$}
    
    \For{j = 0 \textbf{to} Num}
    \State Signal $\gets$ T[j] \Comment{Single Signal in the data} 
    \State Sequence $\gets$ \Call{Mapping}{Signal, RangeBounds, A} 
    \State Seqkmers $\gets$ \Call{Computekmers}{Sequence}
    \State L = Length(Seqkmers)

        \State \texttt{$kmersCount \gets [0] * \vert A \vert^{kmersize}$}
        
        \For{$kmer$ in Seqkmers} 
            \State \texttt{$idx = Globalkmers.index(kmer)$} 
            \State \texttt{$kmersCount[idx] \gets kmersCount[idx] + 1$} 
    
        \EndFor    
        \State \texttt{$\phi.append(kmersCount)$}
    \EndFor

\State \Return $\phi$
\EndFunction
\end{algorithmic}
\caption{Sequence Embedding generation from Time Series Data}
\label{algo_embedding}
\end{algorithm}

\begin{algorithm}[h!]
\begin{algorithmic}[1]
\Statex \textbf{Input: }\texttt{Flattened Time Series Data ($F$)}
\Statex \textbf{Output: }\texttt{Boundaries of Ranges ($B$)}
\Function{ComputeRanges}{$F$}
\State MaxValue$\gets$ Max(F) \Comment{Finding the maximum value}
\State MinValue$\gets$ Min(F) \Comment{Finding the minimum value}
\State $NumRanges$ $\gets$ 26
\Comment{Number of ranges}
\State d $\gets$ $\frac{MaxValue- MinValue}{NumRanges}$
\Comment{The interval in a range }
\State B $\gets$ [] \Comment{Initialize an empty list} 
    \For{i = 0 \textbf{to} $NumRanges$ + 1}
    \If{i=0} 
    \State Bound = F[i]
    \Else 
    \State Bound = Bound + d    
    \Comment{Compute bounds of ranges}
    \EndIf
    \State B.append(Bound)
    \EndFor
\State \Return $B$
\EndFunction
\end{algorithmic}
\caption{Computing Ranges}
\label{algo_range}
\end{algorithm}

\begin{algorithm}[h!]
\begin{algorithmic}[1]
\Statex \textbf{Input: }\texttt{Time Series Signal($\alpha$), RangeBounds (R), Alphabets (A)}
\Statex \textbf{Output: }\texttt{Character Sequence ($S$)}
\Function{Mapping}{$\alpha$, R, A}
    \State NumValues = Length($\alpha$) \Comment{Number of Signal points}
    \State S $\gets$ [] \Comment{Initialize an empty list} 
        \For{m = 0 \textbf{to} NumValues} 
        \State Val = Signal[m] \Comment{Value of one signal point} 
            \For{i=0 \textbf{to} NumRanges}
            \If{R[i] <= Val < R[i+1]}
            \State Character = A[i]
            \EndIf
            \EndFor
        \State S.append(Character)    
        \EndFor
\State \Return $S$
\EndFunction
\end{algorithmic}
\caption{Converting Time Series Data to Alphabetical Representation} 
\label{algo_conversion}
\end{algorithm}

\begin{algorithm}[h!]
\begin{algorithmic}[1]
\Statex \textbf{Input: }\texttt{ Character Sequence (C), ksize}
\Statex \textbf{Output: }\texttt{ List of $k$-mers in a sequence ($K$)}
\Function{Computekmers}{C, ksize}

 \State K $\gets$ [] \Comment{Initialize an empty list} 
    \State LenSeq $\gets$ Length(C)
        \For{i = 0 \textbf{to} LenSeq - ksize + 1}
        \State kmer = C[i:i+ksize] \Comment{Extract $k$-mers from sequence}
        \State K.append(kmer)
        \EndFor

\State \Return $K$
\EndFunction
\end{algorithmic}
\caption{Computing $k$-mers} 
\label{algo_kmers}
\end{algorithm}

\section{Experimental Setup}\label{sec_ES}
We employed a standard 60-10-30\% training, validation, and test set for our experiments, reporting the mean results across 100 iterations. To evaluate the classifiers' (i.e. Support vector machines(SVM), Naive Bayes (NB), Multilayer perceptron (MLP), K-nearest neighbors (KNN), Random Forest (RF), Logistic regression (LR),  and Decision tree (DT)) performance, we utilized various metrics, including Average Accuracy, Precision, Recall, F1 (weighted), F1 (Macro), F1 (Micro), ROC AUC (one-vs-rest), and training runtime. As a baseline, we use the method proposed in~\cite{ali2023information} (i.e. a feature engineering-based method), along with Inception Time~\cite{ismail2020inceptiontime} (a method based on Inception-v4 architecture, comprised of an ensemble
of deep Convolutional Neural Network model), Time Series Transformer (TST)~\cite{zerveas2021transformer} (a transformer-based model for the embedding generation of time series data), and Time Series (TS) Sequencer~\cite{tatsunami2022sequencer} (a deep LSTM-based model for end-to-end learning).


For the dataset, we used the human activity smartphone sensor data from the accelerometer, magnetometer, and gyroscope as proposed in~\cite{ali2023information}. The dataset is originally collected from $29$ users performing different human activities on their smartphones. The total number of data points (i.e. time series signals) is $112$. As the target variable, we use the Gender of the users, the hand in which users were holding smartphones while performing different tasks, the application they were using on the smartphone, and the age of the users. 
The unique values for the age label of the $29$ users are $<20$, 20-25, 25-30, 30-35, $>35$ while their respective count is 2, 7, 7, 6, and 7, respectively. 
The unique values for gender label are Male and Female while their respective distribution are 17 and 12, respectively. 
The unique values for the application label are Facebook, Instagram, WhatsApp, and Twitter while their distributions are 28, 29, 28, and 27, respectively.
The unique values for the Hand label are Left Handed, Right Handed, and Both hands (i.e. holding the smartphone with both hands) while their distributions are 10, 12, and 7, respectively.

\section{Results And Discussion}\label{sec_RD}

The results for the age prediction are shown in Table~\ref{tbl_age_results} for the baseline and our proposed method. We can observe that the proposed method outperforms the baseline for all metrics other than the training runtime. For the training runtime, since the embedding size of the baseline is smaller due to the usage of a smaller set of statistical features, their method can help ML models to train faster. However, in terms of predictive performance, the proposed method outperforms the baseline by $2.8\%$ in terms of average accuracy and $1.9\%$ in terms of ROC-AUC.

\begin{table}[h!]
  \centering
  \caption{Classification results (Averaged over 100 runs) on age prediction. The best values are shown in bold.}
 \resizebox{0.99\linewidth}{!}{
  \begin{tabular}{cccccccc | p{1.3cm}}
    \toprule
    \multirow{2}{1.5cm}{Method} & \multirow{2}{0.7cm}{ML Algo.} & \multirow{2}{*}{Acc.} & \multirow{2}{*}{Prec.} & \multirow{2}{*}{Recall} & \multirow{2}{0.9cm}{F1 weigh.} & \multirow{2}{0.9cm}{F1 Macro} & \multirow{2}{1.2cm}{ROC- AUC} & Train. runtime (sec.) \\	
    \midrule	\midrule	

    \multirow{1}{3cm}{Inception Time~\cite{ismail2020inceptiontime}} 
& - & 0.264 & 0.103 & 0.264 & 0.144 & 0.129 & 0.507 & 0.124 \\
\midrule
\multirow{1}{3cm}{TST~\cite{zerveas2021transformer}} 
& - & 0.382 & 0.207 & 0.382 & 0.264 & 0.2 & 0.550 & 0.201 \\
\midrule
\multirow{1}{3cm}{TS Sequencer~\cite{tatsunami2022sequencer}} 
& - & 0.411 & 0.392 & 0.411 & 0.335 & 0.283 & 0.5902 & 0.247 \\
\midrule
\multirow{7}{3cm}{Feature Engineering~\cite{ali2023information}} 
& SVM & 0.917 & 0.931 & 0.917 & 0.916 & 0.893 & 0.940 & 0.014 \\
& NB & 0.736 & 0.779 & 0.736 & 0.730 & 0.722 & 0.827 & 0.015 \\
& MLP & 0.925 & 0.938 & 0.925 & 0.923 & 0.905 & 0.948 & 0.227 \\
& KNN & 0.843 & 0.836 & 0.843 & 0.825 & 0.752 & 0.867 & 0.016 \\
& RF & 0.915 & 0.924 & 0.915 & 0.911 & 0.884 & 0.934 & 0.139 \\
& LR & 0.911 & 0.923 & 0.911 & 0.907 & 0.882 & 0.933 & 0.020 \\
& DT & 0.837 & 0.856 & 0.837 & 0.833 & 0.810 & 0.891 & \textbf{0.009} \\
\midrule

\multirow{7}{0.7cm}{Ours} 
& SVM & 0.871 & 0.900 & 0.871 & 0.871 & 0.852 & 0.909 & 0.017 \\
 & NB & 0.841 & 0.899 & 0.841 & 0.845 & 0.853 & 0.903 & 0.012 \\
 & MLP & 0.894 & 0.917 & 0.894 & 0.894 & 0.892 & 0.937 & 0.318 \\
 & KNN & 0.547 & 0.633 & 0.547 & 0.538 & 0.534 & 0.720 & 0.013 \\
 & RF & 0.894 & 0.909 & 0.894 & 0.891 & 0.871 & 0.913 & 0.0127 \\
 & LR & \textbf{0.953} & \textbf{0.958} & \textbf{0.953} & \textbf{0.953} & \textbf{0.949} & \textbf{0.967} & 0.033 \\
 & DT & 0.888 & 0.904 & 0.888 & 0.886 & 0.821 & 0.899 & 0.010 \\

    \bottomrule
  \end{tabular}
  }
  \label{tbl_age_results}
\end{table}

For the gender attribute, the maximum average accuracy and ROC-AUC that we got for the baseline are $0.982$ and $0.980$ using the SVM classifier. However, our proposed method's highest respective values are $0.990$ and $0.991$, which we got using the logistic regression classifier.
Similarly, for the hand attribute, the maximum average accuracy and ROC-AUC that we got for the baseline are $0.434$ and $0.563$ using the decision tree classifier. However, our proposed method's highest respective values are $0.445$ and $0.572$, which we got using the decision tree classifier.
Moreover, for the application attribute, the maximum average accuracy and ROC-AUC that we got for the baseline are $0.521$ and $0.688$ using the multi-layer perceptron classifier. However, our proposed method's highest respective values are $0.533$ and $0.695$, which we got using the logistic regression classifier. Note that we have not reported detailed results in tables for Gender, Application, and Hand attributes as we did for the Age attribute in Table~\ref{tbl_age_results} due to space limitations. Therefore, we only reported the best results for the baseline and proposed method for fair comparison. Moreover, it is noted that the proposed method consistently outperformed the baseline for all attributes in terms of all evaluation metrics other than classifier training runtime.

Compared to the deep learning and transformer-based baselines, i.e. Inception Time~\cite{ismail2020inceptiontime}, TST~\cite{zerveas2021transformer}, and TS Sequencer~\cite{tatsunami2022sequencer}, we can observe that the proposed method significantly outperform all these baselines for all evaluation metrics.
This behavior is primarily due to the data-hungry nature of deep learning models, which require large amounts of labeled data to perform effectively~\cite{ismail2019deep}. Many real-world time series datasets, particularly in specialized or emerging fields, often consist of limited data, making deep learning approaches less practical. Additionally, deep learning models tend to be computationally intensive, demanding significant resources for training and deployment~\cite{ismail2019deep,ismail2020inceptiontime}. By focusing on symbolic and bioinformatics-inspired methods, our method aims to develop more accessible and resource-efficient techniques for time series analysis and classification, hence showing higher predictive performance compared to all baselines.

Since we are running our experiments $100$ times and reporting the average results, we also noted the standard deviations (SD) of $100$ runs to observe the stability of the computed results. We noted that the SD values were very low in the majority of the cases, i.e. $<0.02$, which showed that there is not much variation in the reported results. Moreover, we used the famous student t-test to evaluate the statistical significance of the classification results. Since the SD values were very low, the p-values were also $<0.05$, hence showing that the reported results are statistically significant.

\section{Conclusion}\label{sec_conclusion}
This paper presents a novel method for transforming time series signals into biological sequence-type representations using alphabetic mapping and $k$-mer strategy. 
We generated character sequences that retain the temporal ordering and relative magnitudes of the original data. This transformation enables the application of advanced sequence analysis techniques from the field of bioinformatics to time series data.
Our experimental results demonstrate that the proposed method enhances classification accuracy. 
Future work could look into transformation for facilitating transfer learning from the bioinformatics domain, where powerful representation learning techniques, such as those used for DNA and protein sequence analysis, can be repurposed to improve time series analysis. 



\bibliographystyle{IEEEtran}
\bibliography{references}

\end{document}